\def\year{2020}\relax
\documentclass[letterpaper]{article} 
\usepackage{aaai20}  
\usepackage{times}  
\usepackage{helvet} 
\usepackage{courier}  
\usepackage[hyphens]{url}  
\usepackage{graphicx} 
\usepackage{graphicx}  
\usepackage{amsfonts}
\usepackage{amsmath}
\usepackage{bm}
\usepackage{subfig}
\usepackage{multirow}
\DeclareMathOperator*{\argmax}{argmax}
\newcommand{\ie}{\textit{i.e.}}

\title{RDSNet: A New Deep Architecture for\\Reciprocal Object Detection and Instance Segmentation}

\author{Shaoru Wang,\textsuperscript{\rm 1,4}\thanks{The work was done when Shaoru Wang was an intern in Horizon Robotics Inc.} Yongchao Gong,\textsuperscript{\rm 2} Junliang Xing,\textsuperscript{\rm 1} Lichao Huang,\textsuperscript{\rm 2} \\ 
\Large \textbf{Chang Huang,\textsuperscript{\rm 2} Weiming Hu\textsuperscript{\rm 1,3,4}} \\
\textsuperscript{\rm 1}National Laboratory of Pattern Recognition, Institute of Automation, Chinese Academy of Sciences\\ 
\textsuperscript{\rm 2}Horizon Robotics Inc.
\textsuperscript{\rm 3}CAS Center for Excellence in Brain Science and Intelligence Technology\\
\textsuperscript{\rm 4}University of Chinese Academy of Sciences\\
wangshaoru2018@ia.ac.cn, \{jlxing,wmhu\}@nlpr.ia.ac.cn, \{yongchao.gong,lichao.huang,chang.huang\}@horizon.ai
}

\begin{document}
\maketitle
\begin{abstract}
Object detection and instance segmentation are two fundamental computer vision tasks. They are closely correlated but their relations have not yet been fully explored in most previous work. This paper presents RDSNet, a novel deep architecture for reciprocal object detection and instance segmentation. To reciprocate these two tasks, we design a two-stream structure to learn features on both the object level (\ie, bounding boxes) and the pixel level (\ie, instance masks) jointly. Within this structure, information from the two streams is fused alternately, namely information on the object level introduces the awareness of instance and translation variance to the pixel level, and information on the pixel level refines the localization accuracy of objects on the object level in return. Specifically, a correlation module and a cropping module are proposed to yield instance masks, as well as a mask based boundary refinement module for more accurate bounding boxes. Extensive experimental analyses and comparisons on the COCO dataset demonstrate the effectiveness and efficiency of RDSNet. The source code is available at \url{https://github.com/wangsr126/RDSNet}.
\end{abstract}

\section{Introduction}
Object detection and instance segmentation are two fundamental and closely related tasks in computer vision, focusing on progressive image understanding on the object level and the pixel level respectively. Due to the application of deep neural networks, recent years have witnessed significant advances of these two tasks. However, their relations have not yet been fully explored in most previous work. Therefore, it remains meaningful and challenging to improve the performance of these two tasks by leveraging the interaction between the object-level and pixel-level information.

\begin{figure}[t]
	\centering
	\includegraphics[width=1\linewidth]{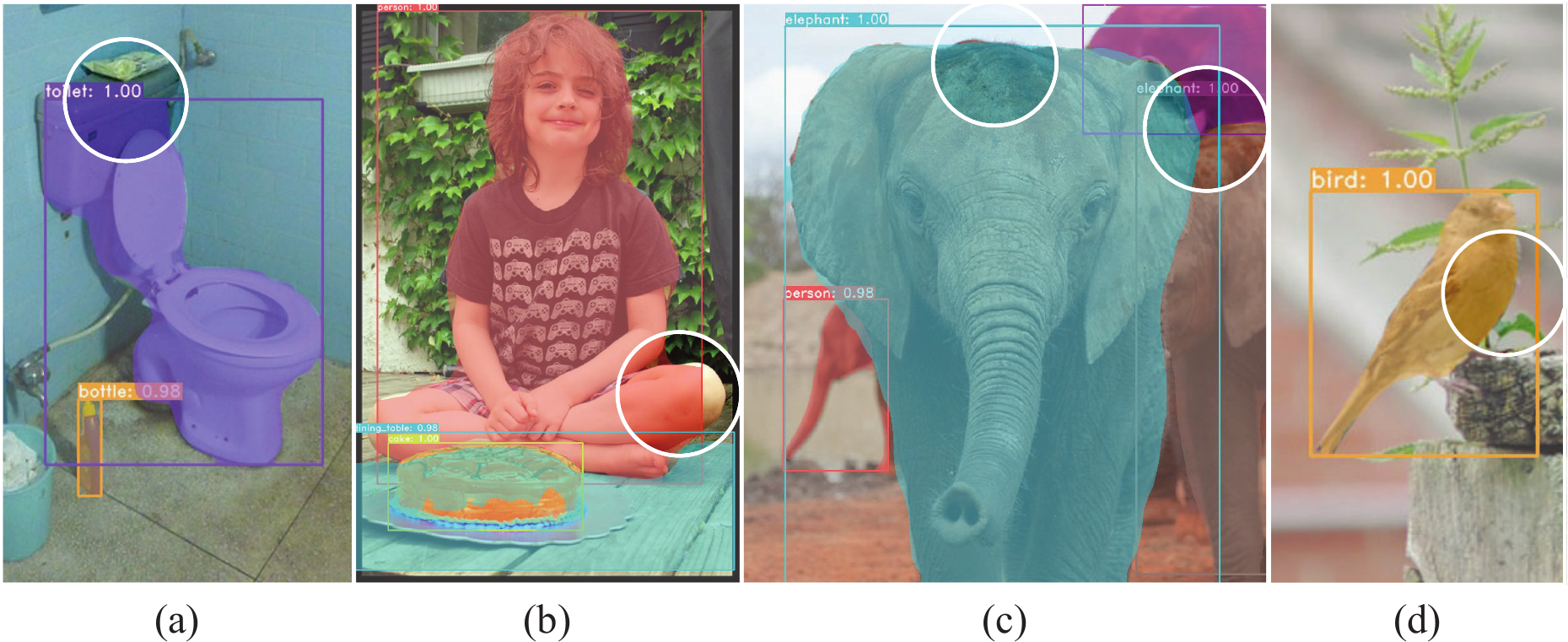}
	\caption{Localization errors in object detection. (a)(b) Boxes do not fully enclose objects. (c)(d) Boxes do not enclose objects tightly. Most of these errors can be easily corrected if we fully leverage the reciprocal relations between the object detection and instance segmentation tasks. Results are obtained by Mask R-CNN \cite{he2017mask}.}
	\label{fig:error}
\end{figure}

The goal of object detection is to localize each object with a rectangular bounding box and classify it into a specific category. In this task, one of the most critical challenges lies in object localization, namely the specification of an inclusive and tight bounding box. As can be commonly observed in many state-of-the-art methods, localization error can easily degrade their performance, as illustrated in Fig.~\ref{fig:error}. Localization error mainly results from the mechanism of using regression method to obtain the bounding boxes, as point-wise regression is not directly aware of the whole object. For this reason, it is more rational to cast object localization into a pixel-level task, which is consistent with the definition of bounding box, \ie, the minimum enclosing rectangle of the object mask. Therefore, if the object masks are provided, it will be more straightforward and accurate to obtain the bounding boxes according to the masks.

Instance segmentation aims to further predict the per-pixel binary mask of each object besides category. The core idea of instance segmentation is to introduce instance-aware pixel category. Currently, most existing approaches follow a two-stage  paradigm (\textit{e.g.} Mask R-CNN \cite{he2017mask}), that is, masks are generated separately for each detection proposal. In this way, the masks are aware of individual object instances naturally. However, such step-by-step process makes the masks heavily depend on the bounding boxes obtained by the detector and vulnerable to their localization errors. Moreover, the utilization of the operator as ROI pooling \cite{girshick2015fast} largely restricts the mask resolution for large objects. The FCIS model \cite{li2017fully} introduces position-sensitive map for instance-aware segmentation, but the resulting masks are still restricted to the detection results. Some other methods get rid of detectors \cite{fathi2017semantic} , but they are inferior in accuracy. The origin of these drawbacks mainly lies in the insufficient and inadequate utilization of the object-level information.

According to the above analyses, object detection and instance segmentation have non-negligible potentials to benefit from each other. Unfortunately, few existing works focus on the relations between them. HTC \cite{chen2019hybrid} is a representative work that adopts cascade architecture for progressive refinement of the two tasks and achieves promising results. However, such multi-stage design brings relatively high computation cost.

In this work, we present a \textbf{R}eciprocal Object \textbf{D}etection and Instance \textbf{S}egmentation \textbf{Net}work (RDSNet) to leverage the relation between these two tasks. RDSNet adopts a two-stream structure, \ie, object stream and pixel stream. Features from such two streams are extracted jointly and simultaneously, and then fused alternately across each other. Specifically, the object stream focuses on the object-level features and is formed by  a regression-based detector, while the pixel stream focuses on the pixel-level features and follows the FCN \cite{long2015fully} architecture to ensure high-resolution outputs. To leverage object-level cues from the object stream, a correlation module and a cropping module are proposed, which introduce the awareness of instances and the translation-variance property to the pixel stream, and yield instance-aware segmentation masks. In turn, a mask based boundary refinement module is proposed to alleviate the localization error with the help of the pixel stream, \ie, produce more accurate bounding boxes based on the instance masks.

RDSNet takes a sufficient consideration of reciprocal relations between object detection and instance segmentation tasks.
Compared with previous approaches, it has the following three advantages: 1) Masks generated by RDSNet have consistent high resolution for objects in various scales; 2) Masks are less dependent on detection results thanks to the ingenious cropping module; 3) More accurate and tighter bounding boxes are obtained with a novel pixel-level formulation of object bounding box locations. 

Our main contribution is that we explore the reciprocal relations between the object detection and instance segmentation tasks. And an end-to-end unified architecture RDSNet is proposed to leverage such object-level and pixel-level tasks from each other, which demonstrates the potentials of the multi-tasks fusion concept. 

\section{Related Works}
\subsubsection{Object Detection.}
Most modern CNN-based detectors rely on the regression methods to obtain the bounding boxes of objects. One typical approach is the anchor based methods \shortcite{dai2016r,liu2016ssd,lin2017feature,lin2017focal}, which is first used in the Faster R-CNN model \cite{ren2015faster}. Dense anchors of multiple scales and aspect ratios are put at each sliding-window location and serve as regression references. Detectors classify such anchor boxes and regress the offsets from the anchor boxes to the bounding boxes. Another branch of regression-based detectors eliminates the anchor boxes, \ie, anchor-free, which directly predict the center of objects and regress boundary\footnote{Unless otherwise specified, we use \texttt{boundary} to refer to the box boundary, not object boundary. } at each location \cite{DBLP:journals/corr/HuangYDY15,Yang2019RepPoints,Tian2019FCOS}. In this work, we propose a simple but effective method, extending above regression-based detectors to the instance segmentation task and the localization accuracy will be improved.

Recently, some newly proposed approaches detect objects as keypoints related to the bounding box \cite{law2018cornernet,zhou2019bottom,duan2019centernet}, but complicated post-processing is required to group such points belonging to the same instance. 

\subsubsection{Instance Segmentation.}

\begin{figure*}[ht]
	\centering
	\includegraphics[width=1\linewidth]{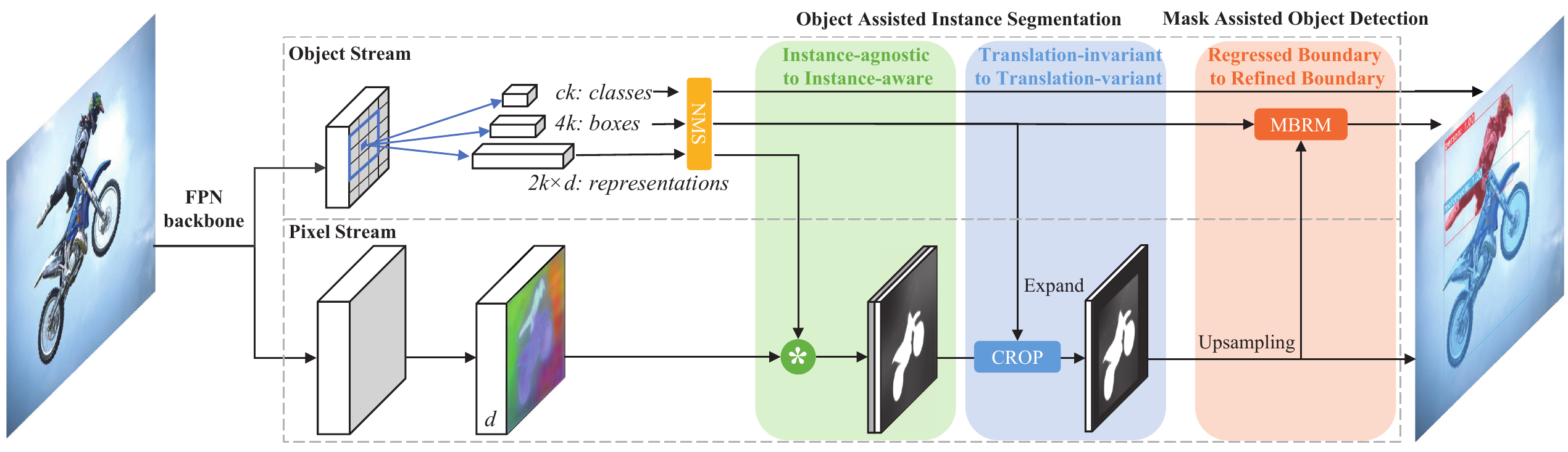}
	\caption{The architecture of the proposed RDSNet, which follows a two-stream structure, \textit{i.e.}, object stream and pixel stream. Information from these two streams are mutually interacted by several well-designed modules: the correlation module and cropping module introduce the awareness of instance and the translation variance to pixel stream, assisting in generating the instance masks (see Sec.~\ref{sec:o2p}). In turn, the instance masks assist the object stream in obtaining more accurate bounding boxes (see Sec.~\ref{sec:p2o}). \textit{c} denotes the class number, $k$ denotes the anchors number at one location, $d$ denotes the representation dimensions, and $\star$ denotes the convolution operation.}
	\label{fig:pipeline}
\end{figure*}

Existing instance segmentation approaches can be grouped as two-stage and one-stage ones. Two-stage approaches follows the top-down process, \ie, \textit{detect-then-segment}\cite{he2017mask}, in which the object is firstly detected as a bounding box and then a binary mask is generated for each object. Approaches built on Mask R-CNN (\textit{e.g.} \cite{liu2018path}) have dominated several popular benchmarks \cite{lin2014microsoft,cordts2016cityscapes}. However, such step-by-step process makes mask quality heavily depend on the box accuracy.

One-stage approaches are also known as single-shot ones, since objects are directly classified, located and segmented without generating candidate region proposals. A branch of one-stage approaches \shortcite{arnab2017pixelwise,liu2017sgn,fathi2017semantic,de2017semantic,neven2019instance} follows the bottom-up process, \ie, \textit{label-pixels-then-cluster}, in which pixels are firstly labeled with a category or embedded to a feature space, and then grouped into each object. These approaches derive from methods developed for semantic segmentation, and higher-resolution masks are obtained naturally. However, the unawareness of objects status (numbers, locations, \textit{etc.}) beforehand complicates the design of predefined categories or embedded space, resulting in inferior results. We consider the origin of predicament lies in the lack of object-level information. Another branch of one-stage approaches \cite{li2017fully,dan2019yolact} is proposed to leverage the top-down and bottom-up approaches jointly. These methods follow the \textit{label-pixels-then-cluster} process roughly while the grouping method relies on detection results, directly or indirectly (\textit{e.g.}, cropping the masks with bounding boxes predicted by the detector). Our approach follows this process in general, but the object-level information are introduced to simplify the embedded space design with a correlation module, and a modified cropping module is proposed to lower the dependencies of the instance masks on the bounding boxes. 

\subsubsection{Boundary Refinement}
Cascade R-CNN \cite{cai2018cascade} adopts a cascade architecture to refine the detection results by multi-stage iterative localization. HTC \cite{chen2019hybrid} further improves the information flow. But these methods are designed for two-stage approaches. Instead, our method refines the boundary localization based on a novel formulation, with the compatibility to one-stage approaches and less computation.

\section{RDSNet}
In this section, we first introduce the overall architecture of RDSNet, where the core is a two-stream structure, consisting of the object stream and the pixel stream, as shown in Fig.~\ref{fig:pipeline}. Then the bidirectional interaction across the two streams are presented, \ie, leveraging the object-level information to facilitate instance segmentation and the pixel-level information to facilitate object detection.

\subsection{Two-stream structure}
The core of RDSNet is the two-stream structure, namely the object stream and the pixel stream. These two streams share the same FPN \cite{lin2017feature} backbone and are then separated for each corresponding task. Such parallel structure enables the decoupling of object-level and pixel-level information and alterable resolutions for different tasks. 

\subsubsection{Object Stream.}
The object stream focuses on object-level information, including object categories, locations, \textit{etc.} It can be formed by various regression-based detectors \cite{liu2016ssd,redmon1804yolov3,lin2017focal}. In addition, we add a new branch in parallel with the classification and regression branches to extract the object feature for each anchor (or location). This stream is responsible for producing detection results that will later be refined by pixel-level information (see Sec.~\ref{sec:p2o}).

\subsubsection{Pixel Stream.}
The pixel stream focuses on pixel-level information, and follows the FCN \shortcite{long2015fully} design for high-resolution outputs. Specifically, per-pixel features are extracted in this stream, and then used to generate instance masks by utilizing object-level information (see Sec.~\ref{sec:o2p}).

\subsection{Object Assisted Instance Segmentation\label{sec:o2p}}
This subsection introduces a novel way to yield instance masks by leveraging the object-level information with new designed correlation and cropping modules.

\subsubsection{From Instance-agnostic to Instance-aware.}
Instance segmentation aims to assign an instance-aware category to each pixel, but it often suffers from the ambiguity that no predefined categories for pixels are available due to the uncertain numbers and locations of objects in 2D image plane. A proper solution is to leverage the object-level information to introduce the awareness of instances. To this end, a correlation module is designed to link each pixel to its corresponding instance according to the similarity between their representations, which are learned from the object stream and the pixel stream, respectively.

Given an object $o$, we denote its representation by $\phi(v_o) \in \mathbb{R}^{2\times d\times 1\times 1}$, where $v_o$ represents the feature of that object from object stream, and $d$ is the dimension of the representation. The $2$ dimension of $\phi(v_o)$ indicates that we take both foreground and background into consideration. Similarly, we denote the pixel representations of the entire image as $\Psi(U) \in \mathbb{R}^{1\times d\times h_f\times w_f}$, where $U$ represents the feature map from pixel stream, $h_f$ and $w_f$ are the spatial dimensions of $\Psi(U)$.

\begin{figure}[t]
	\centering
	\includegraphics[width=1\linewidth]{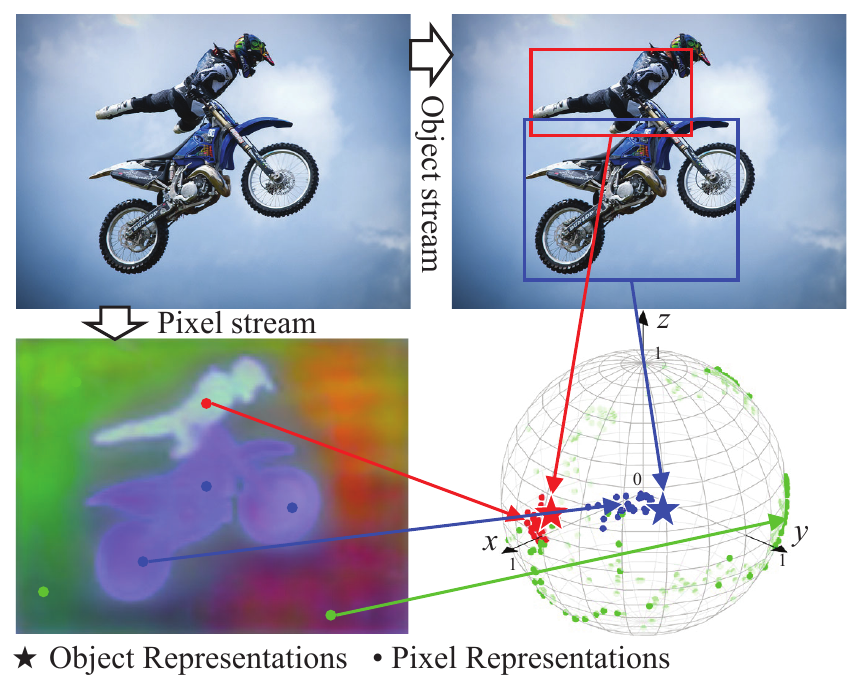}
	\caption{Illustration for the representations of the objects\footnotemark\, and pixels, both of which are embedded into $d$-dimension feature space in the object and pixel streams respectively. Pixel representations are close to corresponding instance representation in feature space and different objects have distinct representations. Dimension reduction (from $d$ to 3) and L2 normalization are performed to representations.}
	\label{fig:metric}
\end{figure}

The purpose of the correlation module is to measure the similarity between $\phi(v_o)$ and $\Psi(U)$. The correlation operation is defined by 
\begin{equation}
\label{eq:corr}
    M_o=\mathrm{softmax}(\Psi(U) \star \phi(v_o))\;,
\end{equation}
where $\star$ stands for the convolution operator. The two channels of similarity map $M_o \in \mathbb{R}^{2 \times 1\times h_f\times w_f}$ can be viewed as the foreground and background probabilities of each pixel corresponding to object $o$. Pixel-wise cross entropy loss is appended on $M_o$ in training stage. For all objects in an image, the correlation operation is repeated respectively and synchronously. The correlation module enables the mask generator to be trained end-to-end. In a sense, the training process of our approach with correlation is similar to metric learning \cite{fathi2017semantic}, that is, pulling the representations of foreground pixels towards their corresponding object representation in feature space, and pushing those of background pixels away, as illustrated in Fig.~\ref{fig:metric}.

\footnotetext{Only foreground channel is taken.}

\subsubsection{From Translation-invariant to Translation-variant.}

Unlike most two-stage instance segmentation approaches \cite{he2017mask}, mask generated by the above correlation module for each object covers the whole image, regardless of the object size and location.  Such characteristic guarantees high-resolution results, but noise is easily involved.  This drawback is largely attributed to the translation-invariant property of convolution: any two pixels with similar appearances tend to have similar representations, although they might actually belong to different instances or background. The property makes it difficult to exclude the noise directly due to the absence of spatial information in the pixel representations. Fortunately, we can overcome this drawback simply by using the bounding boxes produced by the object stream as they can provide adequate spatial restrictions. Specifically, for each object, pixels outside its bounding box are directly set as background and ignored during training. Such cropping strategy makes the instance mask restricted to the internal area of the bounding box and the pixels faraway are not involved in the instance mask even though they have similar appearances. However, simply cropping with such bounding boxes makes the instance masks suffer from the localization errors of detection results (as shown in Fig.~\ref{fig:error} (a)(b)) and unexpectedly leads to a strong coupling relation between detection and segmentation results.

To address this issue, a compromise is made by cropping the masks with expanded bounding boxes. During inference, such strategy guarantees relatively low dependencies of masks on bounding boxes, and the pixels far enough are not involved in the masks. Moreover, cropping with the expanded bounding boxes enables a reasonable diversity of negative pixels during training. Two extreme cases, \ie, no cropping and cropping without expanding, are both harmful for our task because too much diversity causes convergence difficulty while insufficient diversity results in deficient feature space, respectively.

It should be noted that cropping with expanded bounding boxes makes more background pixels involved for each object during training, making the background pixels easily dominate the training procedure. To maintain a manageable balance between the foreground and background (1:1 in our experiments), online hard example mining (OHEM) \cite{shrivastava2016training}  for background pixels is adopted.

\subsection{Mask Assisted Object Detection}\label{sec:p2o}
In this section, we introduce how to enhance the detection results by utilizing the pixel-level information. According to the aforementioned analyses, pixel-level information has the potential to benefit the detection task, especially for object boundary localization. To this end, we develop a new formulation for boundary localization based on the Bayes' theorem. In this formulation, we comprehensively utilize the bounding box and instance mask obtained from the object stream and pixel stream to get a more accurate bounding box of each object. Based on this formulation, a mask based boundary refinement module (MBRM) is proposed.

\begin{table*}[ht]
    \centering
    \resizebox{1\linewidth}{!}{
    \begin{tabular}{l|c|c|c|c|c c c|c c c}
    \hline
        Method & Scale & epoch & aug & FPS & AP$^{m}$ & AP$_{50}^{m}$ & AP$_{75}^{m}$ & AP$_S^{m}$ & AP$_M^{m}$ & AP$_L^{m}$\\
        \hline
        \textit{two-stage:}& & & & & & & & & & \\
        Mask R-CNN \cite{he2017mask} $\dag$ & 800 & 12 & $\circ$ & \textbf{9.5}/V & 36.2 & \textbf{58.3} & 38.6 & 16.7 & 38.8 & 51.5\\
        MS R-CNN \cite{huang2019mask} $\dag$ & 800 & 12 & $\circ$ & 9.1/V & \textbf{37.4} & 57.9 & \textbf{40.4} & \textbf{17.3} & \textbf{39.5} & \textbf{53.0}\\
        \hline
        \hline
        \textit{one-stage:}& & & & & & & & & & \\
        FCIS \cite{li2017fully} & 600 & 12 & $\circ$ & 6.6/P & 29.2 & 49.5 & - & 7.1 & 31.3 & 50.0\\
        YOLACT \cite{dan2019yolact} & 550 & 48 & $\surd$ & \textbf{33.0}/P & 29.8 & 48.5 & 31.2 & 9.9 & 31.3 & 47.7\\
        RDSNet$_s$ (ours) & 550 & 48 & $\surd$ & 32.0/P & \textbf{32.1} & \textbf{53.0} & \textbf{33.4} & \textbf{11.0} & \textbf{33.8} & \textbf{51.0}\\
        \hline
        TensorMask \cite{chen2019tensormask} & 800 & 72 & $\surd$ & 2.6/V & \textbf{37.3} & \textbf{59.5} & \textbf{39.5} & \textbf{17.5} & 39.3 & \textbf{51.6}\\
        RDSNet (ours) & 800 & 12 & $\circ$ & \textbf{8.8}/V & 34.6 & 55.8 & 36.7 & 14.9 & 37.4 & 50.3\\
        RDSNet (ours) & 800 & 24 & $\surd$ & \textbf{8.8}/V & 36.4 & 57.9 & 39.0 & 16.4 & \textbf{39.5} & \textbf{51.6}\\
        \hline
    \end{tabular}
    }
    \caption{Instance segmentation results on COCO test-dev. P means Titan XP or 1080Ti, and V means Tesla V100. `aug' means data augmentation during training: $\circ$ is trained with only horizontal flipping augmentation and $\surd$ is trained further with scale augmentation.  $\dag$ means this entry is obtained by models provided by mmdetection \cite{chen2019mmdetection}. }
    \label{tab:segmentation}
\end{table*}

\begin{table*}[t]
    \centering
    \resizebox{1\linewidth}{!}{
    \begin{tabular}{l|l|c|c|c|c c c|c c c}
    \hline
       \multicolumn{2}{l|}{ Method} & Scale & Backbone & FPS & AP$^{bb}$ & AP$_{50}^{bb}$ & AP$_{75}^{bb}$ & AP$_S^{bb}$ & AP$_M^{bb}$ & AP$_L^{bb}$\\
        \hline
        \multicolumn{2}{l|}{\textit{two-stage:}}& & & & & & & & & \\
        \multicolumn{2}{l|}{Mask R-CNN \cite{he2017mask}$\dag$} & 800 & R-101 & 9.5/V & 39.7 & 61.6 & 43.2 & 23.0 & 43.2 & 49.7\\
        \multicolumn{2}{l|}{Cascade Mask R-CNN \shortcite{cai2018cascade}$\dag$ } & 800 & R-101 & 6.8/V & 43.1 & 61.5 & 46.9 & 24.0 & 45.9 & 55.4\\
        \multicolumn{2}{l|}{HTC \cite{chen2019hybrid}$\dag$ } & 800 & R-101 & 4.1/V & 45.1 & 64.3 & 49.0 & 25.2 & 48.0 & 58.2\\
        \hline
        \hline
        \multicolumn{2}{l|}{\textit{one-stage:}}& & & & & & & & & \\
        \multicolumn{2}{l|}{YOLOv3\cite{redmon1804yolov3}} & 608 & D-53 & 19.8/P & 33.0 & 57.9 & 34.3 & 18.3 & 35.4 & 41.9\\
        \multicolumn{2}{l|}{RefineDet \cite{zhang2018single}} & 512 & R-101 & 9.1/P & 36.4 & 57.5 & 39.5 & 16.6 & 39.9 & 51.4\\
        \multicolumn{2}{l|}{CornerNet \cite{law2018cornernet}} & 512 & H-104 & 4.4/P & 40.5 & 57.8 & 45.3 & 20.8 & 44.8 & 56.7 \\
        \hline
        \multirow{3}{*}{RDSNet}&RetinaNet \cite{lin2017focal}  &  & & 16.8/V & 36.0 & 55.2 & 38.7 & 17.4 & 39.6 & 49.7 \\
         &+ mask & 600 & R-101 & 15.5/V & 36.1 & 56.7 & 38.5 & 17.3 & 38.9 & 51.3\\
        &+ MBRM  & & & 14.5/V & 37.3 & 56.7 & 39.3 & 17.0 & 40.0 & 54.0\\
        \hline
        &RetinaNet \cite{lin2017focal}  & & & 10.9/V & 38.1 & 58.5 & 40.8 & 21.2 & 41.5 & 48.2\\
        RDSNet&+ mask & 800 & R-101 & 8.8/V & 39.4 & 60.1 & 42.5 & 22.1 & 42.6 & 49.9 \\
        &+ MBRM  & & & 8.5/V & 40.3 & 60.1 & 43.0 & 22.1 & 43.5 & 51.5 \\
        \hline 
    \end{tabular}
    }
\caption{Object detection results on COCO \texttt{test-dev}. We denote the backbone by \texttt{network-depth}, where \texttt{R}, \texttt{D} and \texttt{H} refer to ResNet \cite{he2016deep}, DarkNet \cite{redmon1804yolov3} and Hourglass \cite{newell2016stacked}, respectively.}
    \label{tab:detection}
\end{table*}

\subsubsection{Mask~Based~Boundary~Refinement~Module.}
Bounding box is originally defined as the minimum enclosing rectangle of an object, indicating that it absolutely depends on the region covered by the instance mask. In this sense, it seems indirect to obtain bounding boxes by regression methods that are commonly adopted in existing object detection approaches. Instead, if an instance mask is provided, a quite straightforward solution is to use its minimum enclosing rectangle as the detection result. This is exactly our baseline named \textit{direct}. In this case, the regressed bounding box is only used for mask generation in the pixel stream. 

Although the regressed bounding boxes might contain localization errors, we think they still provide a reasonable prior for the object boundary location to some extent. Therefore, our formulation leverages the detection and segmentation results jointly. Specifically, we view the coordinate of a boundary as a discrete random variable. From the probabilistic perspective, an object boundary location is the \textit{argmax} of the probability of a coordinate where the boundary locates, namely
\begin{equation}
    x=\argmax_{i}P(X=i|M')\;,
\end{equation}
where $X$ is the discrete random variable for the horizontal coordinate of the left boundary, $M'\in \mathbb{R}^{h\times w}$ is the foreground channel of $M$ in Eq.~(\ref{eq:corr}) up-sampled to the input image size, $h\times w$, with all the dimensions of size $1$ removed, and $P(X=i|M')$ denotes the posterior probability given the corresponding instance mask $M'$.

In the following, we only take the derivation for the left boundary as an example, and it can be easily extended to the other boundaries.

Following the Bayes' theorem, we have
\begin{equation}
    P(X=i|M')=\frac{P(X=i)P(M'|X=i)}{\sum_{t=1}^w P(X=t)P(M'|X=t)}\;,
\end{equation}
where $P(X=i)$ and $P(M'|X=i)$ are the corresponding prior and likelihood probabilities.

Assuming that the boundary is only related to the maximum of each row in $M'$, and it only affects its neighboring pixels, the likelihood probability can be defined as
\begin{align}
    P(M'|X=i)&=P(m^x|X=i)\\
    &=P(m^x_{i-s,\dots,i+s}|X=i)\;,
\end{align}
where $m^x_i=\max\limits_{1\leq j\leq h} M_{ij}'$, and $s$ is a hyper-parameter, describing the influence scope of the boundary on its neighboring pixels. Ideally, a pixel on the boundary only affects its two nearest neighboring pixels, \ie, the one outside the bounding box has probability 0 and the other inside has probability 1. In this case, $s=1$. However, the instance mask is not so sharp, making it difficult to provide a proper formulation for $P(m^x_{i-s,\dots,i+s}|X=i)$. Therefore, we approximate it with a one-dimensional convolution with kernel size $2s+1$, followed by a \textit{sigmoid} function for normalization, and the parameters are learned by back-propagation. 

For $P(X=i)$, we simply adopt a discrete Gaussian distribution
\begin{equation}
    P(X=i)=\alpha e^{-(i-\mu)^2/2\sigma_x ^2}\;,
\end{equation}
where $\alpha$ is the normalization coefficient. 
 Obviously, the distribution of the boundary location is related to the instance scale, thus we set
\begin{equation}
    \quad \mu=x_r\;, \;\sigma_x=\gamma w_b \label{equ:sigma}\;,
\end{equation}
where $w_b$ denotes the width of the bounding box, $x_r$ denotes the horizontal coordinate of the regressed left boundary, and $\gamma$ specifies the weighting of the regressed boundary. It can be seen that a smaller $\gamma$ indicates a higher weighting of the regressed boundary, and vice versa.

During training, the ground-truth boundary is transformed to one-hot format along the width or height directions of the image, and cross entropy loss is used to train the above coordinate classification task.

\begin{figure}[t]
    \centering
    \includegraphics[width=1\linewidth]{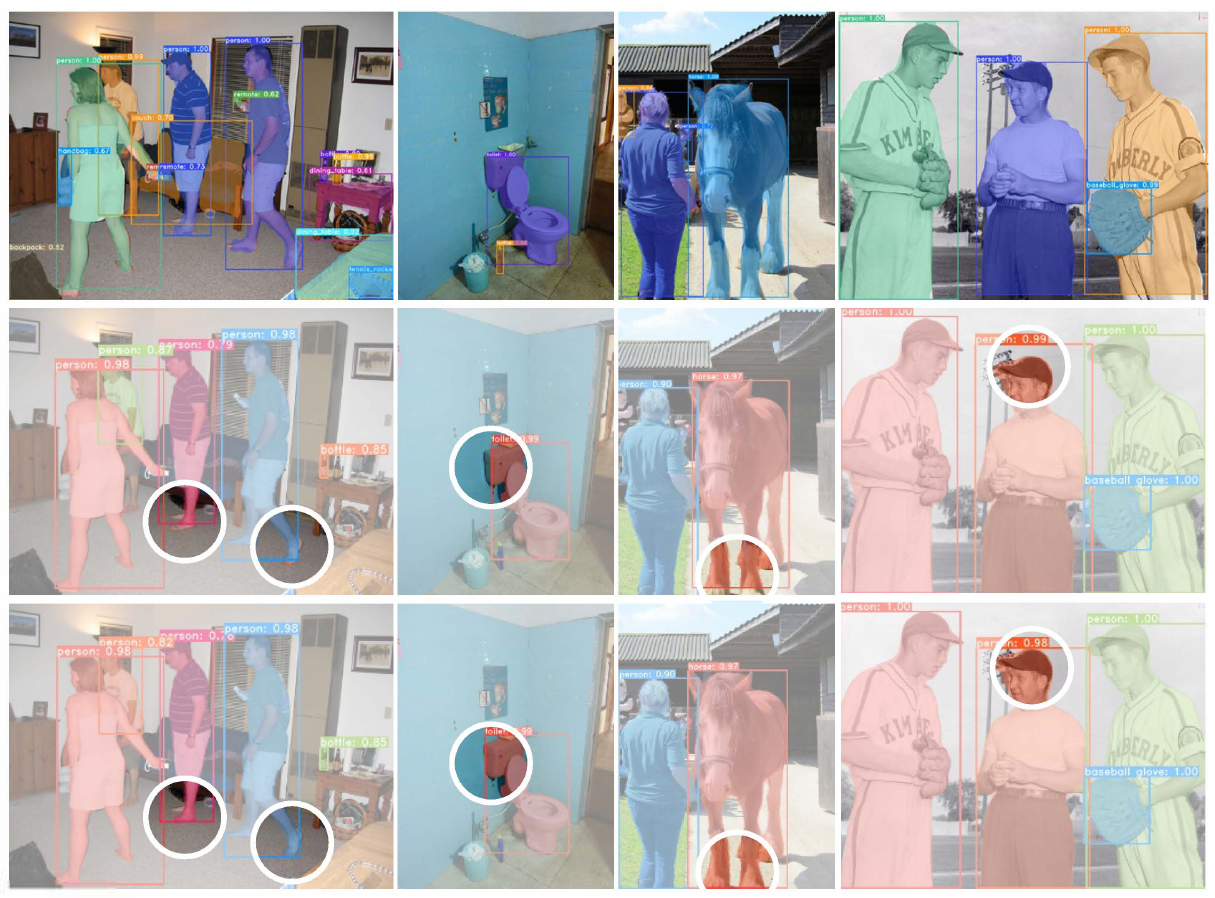}
    \caption{Visual comparisons of some results on COCO \texttt{val2017}. The top, middle and bottom rows are obtained by Mask R-CNN, RDSNet \textit{w/o} expanded cropping or MBRM, and full version of RDSNet. RDSNet gives sharper masks compared to Mask R-CNN. The circled regions highlight the advantage of MBRM in alleviating localization errors.}
    \label{fig:display}
\end{figure}

\subsection{Training and Inference}
Our model is trained with the following multi-task loss:
\begin{equation}
    L=L_{cls}+\lambda_r L_{reg}+\lambda_m L_{mask}\;,
\end{equation}
where $L_{cls}$ and $L_{reg}$ are the commonly used classification and regression losses in detection tasks \cite{ren2015faster,lin2017focal}, and $L_{mask}$ is the pixel-wise cross entropy loss described in Sec.~\ref{sec:o2p}. Only the representations of positive anchors (matched with ground-truth boxes) are fed into the correlation module to generate instance masks, which are then cropped with the expanded ground-truth boxes and used to calculate $L_{mask}$. In other words, pixels outside the expanded boxes are ignored in $L_{mask}$. $L_{refine}$ is the cross entropy loss defined in Sec.~\ref{sec:p2o}. $\lambda_r$ and $\lambda_m$ are hyper-parameters for loss re-weighting. The parameters in MBRM is trained individually with $L_{refine}$ after all others parameters are trained to convergence with $L$. The reason is that MBRM only requires relatively good regression boxes and instance masks.

During inference, the object categories and bounding boxes are first obtained by the detector in the object stream, along with the representation of each instance. Meanwhile, the pixel representations are generated in the pixel stream. Next, only proposals after NMS are processed in the correlation module to generate instance masks, which are then cropped with the expanded boxes obtained by the detector. In order to get the exact coordinates, such instance masks are up-sampled to the input image size and then fed into MBRM. The masks are binarized with threshold $0.4$ at last.

\section{Experiments}
In this section, experimental analyses and comparisons are performed to demonstrate the reciprocal relations between the object detection and instance segmentation tasks. We report the results on COCO dataset \cite{lin2014microsoft} and use the commonly-used metrics for both object detection (AP$^{bb}$) and instance segmentation (AP$^{m}$). We train on \texttt{train2017}, and evaluate on \texttt{val2017} and \texttt{test-dev}.

\subsection{Implementation Details}\label{sec:detail}
We implement RDSNet based on mmdetection \cite{chen2019mmdetection}. We use ResNet-101 \cite{he2016deep} with FPN \cite{lin2017feature} as our backbone. For the object stream, we choose a strong one-stage detector, RetinaNet \cite{lin2017focal} as our detector unless otherwise indicated, as well as our baseline, to validate the effectiveness of our method. 

For the pixel stream, we adopt the architecture of semantic segmentation branch in PanopticFPN \cite{kirillov2019panoptic} to merge the FPN pyramid into a single output, \ie pixel representations, except that the number of channels is modified to 256 for richer representations.

Dimensions of the instance and pixel representations are 32. We use different expanding ratios of bounding boxes for cropping masks during training and inference. During training, we use ground-truth bounding boxes and expand both the heights and the widths of them by $1.5$ times with center point retaining. During inference, the expanding ratio is set to $1.2$. All $\lambda$s are set to $1$.

We train our models on 4 GPUs (2 images per GPU) and adopt $1\times$ training strategy \cite{chen2019mmdetection} along with all other settings same as RetinaNet, and then parameters in MBRM are trained individually for another 1k iterations.

\subsection{Object Assisted Instance Segmentation}

In this section, we first validate the effectiveness of our correlation and cropping module. We compare RDSNet with YOLACT \cite{dan2019yolact}, another one-stage approach for instance segmentation. We adopt the backbone and detection head of YOLACT and apply the correlation module with the expanded cropping strategy (denoted as RDSNet$_s$), compared with the linear combining method with simply cropping in YOLACT. As shown in Tab.~\ref{tab:mask_generator}, 31.0 mAP (\textbf{+1.1} mAP) is achieved for instance segmentation with correlation method compared to 29.9 mAP of YOLACT. What's more, the fast speed is maintained. Compared to only foreground coefficients with additional restriction in YOLACT, modeling both foreground and background representations for each object possibly contributes to easier convergence and thus better results.

Additional ablation experiments in Tab.~\ref{tab:mask_generator} shows the effectiveness of the cropping module. If we simply crop the masks with expanded regressed bounding boxes during inference, performance degradation is observed (row 2 \textit{v.s.} row 3), which indicates that the model is unable to handle the large diversity of background pixels unless expanding strategy is applied during training (row 3 \textit{v.s.} row 5). Once OHEM for negative pixels is adopted, \textbf{1.9} mAP improvement over YOLACT is observed (row 7).

Then, we compare RDSNet with the state-of-the-art methods for instance segmentation. As shown in Tab.~\ref{tab:segmentation}, our method achieves better balance between speed and accuracy among one-stage methods. With small input size (550 or 600), we achieve 32.1 mAP with a real-time speed (32 fps). With 800 input size, RDSNet outperforms most one-stage methods except TensorMask \cite{chen2019tensormask}, which is however nearly 3 times slower. Compared to the two-stage methods, it is worth noting that RDSNet overcomes the inherent drawbacks of Mask R-CNN \cite{he2017mask} to a large extent, such as the low resolution of masks, strong dependencies of masks on bounding boxes, \textit{etc.}, as demonstrated in Fig.~\ref{fig:error} and Fig.~\ref{fig:display} . Besides, we argue that the speed of RDSNet is restricted to the speed of our detector \cite{lin2017focal} (10.9 fps). As shown in Tab.~\ref{tab:detection}, only slight latency is brought to the original detector in RDSNet. As a consequence, further speeding up is achievable by switching to other faster detectors, which is beyond the scope of this work.

\begin{table}[t]
    \centering
        \resizebox{1\linewidth}{!}{
        \begin{tabular}{c|c|c|c c c|c|l}
        \hline
            No. & & Method & TE & OHEM & IE & FPS & AP$^{m}$\\
            \hline
            1 & YOLACT & LC & & & & 33 & 29.9 \\
            \hline
            2 & & & & & & & 31.0$^{+1.1}$  \\
            3 & & & & & $\surd$ & & 30.0 \\
            4 &RDSNet$_s$ & Corr & $\surd$ & & & 32 & 30.7 \\
            5 & & & $\surd$ & & $\surd$ & & 30.8 \\
            6 & & & $\surd$ & $\surd$ & & & 31.6 \\
            7 & & & $\surd$ & $\surd$ & $\surd$ & & \textbf{31.8}$^{+1.9}$ \\
            \hline
        \end{tabular}
        }
    \caption{Demonstration of the effectiveness of the cropping module on COCO \texttt{val2017}. LC: Linear Combination, Corr: Correlation, TE: Expand during training, IE: Expand during inference. Our finally adopted choice (last row) yields the highest mAP. It should be noted that by using Corr instead of LC, RDSNet already outperforms YOLACT by 1.1 in mAP.}
    \label{tab:mask_generator}
\end{table}

\begin{figure}[t]
	\centering
	\includegraphics[width=1\linewidth]{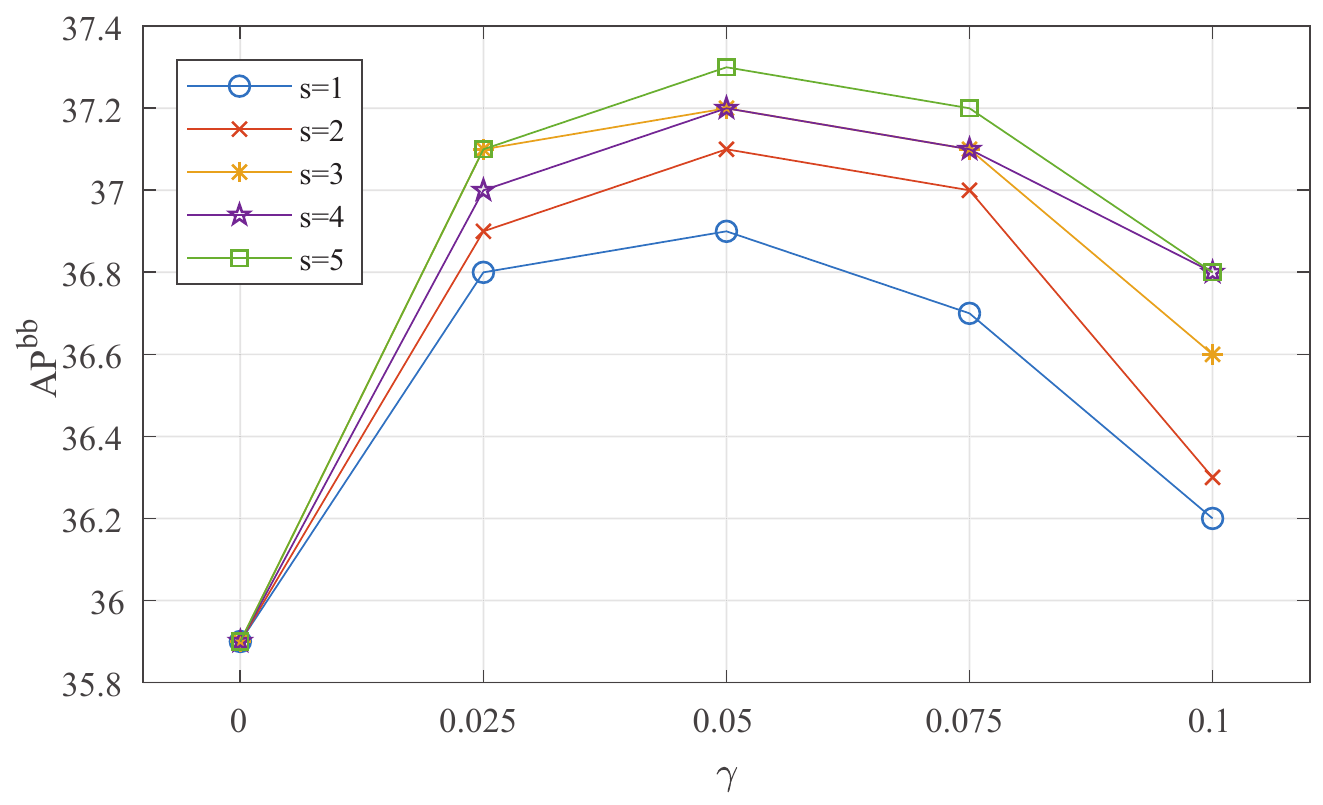}
	\caption{Hyper-parameter sensitivity of MBRM.}
	\label{fig:sigma-k}
\end{figure}

\begin{table}[t]
    \centering
    \resizebox{1\linewidth}{!}{
        \begin{tabular}{c|l|l l l}
        \hline
            Method & $\quad$AP$^{bb}$ & AP$_{S}^{bb}$ & AP$_{M}^{bb}$ & AP$_{L}^{bb}$ \\
            \hline
            baseline & 35.9 & \textbf{17.1} & 39.7 & 53.3 \\
            direct & 34.2$^{-1.7}$ & 11.8$^{-5.3}$ & 37.7$^{-2.0}$ & 55.1$^{+1.8}$\\
            MBRM & \textbf{37.2}$^{+1.3}$ & 16.9$^{-0.2}$ & \textbf{40.8}$^{+1.1}$ & \textbf{56.5}$^{+3.2}$\\
            \hline
        \end{tabular}
        }
    \caption{Demonstration of the effectiveness of MBRM on COCO \texttt{val2017}. Simply regarding the minimum enclosing rectangle of the instance mask as detection results (row 2) does not work well on small objects. However, our MBRM (row 3) works better by introducing the regression bounding box as prior.}
    \label{tab:MBRM}
\end{table}

\subsection{Mask Assisted Object Detection}
For detection task, the key novelty of RDSNet is to refine bounding boxes with instance mask in a one-stage process. As shown in Tab.~\ref{tab:detection}, we find multi-task training with an extra mask generator does bring a certain amount of improvement on our baseline (RetinaNet \cite{lin2017focal}), but further consistent improvement is achieved by MBRM with negligible computational cost. Note that the gain all comes from the more accurate localization of boundary, instead of all other aspects. For fair comparison, only single model results without test-time augmentations are shown in the table.

We further analyze the sensitivity of hyper-parameter in MBRM on COCO \texttt{val2017}, \textit{i.e.}, $s$ and $\gamma$, as shown in Fig.~\ref{fig:sigma-k}. When $\gamma=0$, the refinement module is not activated. We observe that different $\gamma$ results in variant improvement. $\gamma$ around 0.05 works stably so $\gamma=0.05$ is used in all experiments. $s$ indicates how faraway a pixel from the boundary is still affected. Larger $s$ leads to more accurate results within a certain range, while further increasing $s$ does not bring much improvement. We use $s=4$ for all experiments. 

Then, we compare our MBRM with the \textit{direct} method, as shown in Tab.~\ref{tab:MBRM}. We find \textit{direct} method works badly on small objects, which indicates that the prior of regression bounding box is necessary. Our MBRM works better especially for large objects, while slight decline on small objects is negligible, which would be fixed if more precise masks for small objects are provided.

\section{Conclusion}
We have proposed a unified architecture for object detection and instance segmentation, and experimental analyses demonstrate the reciprocal relations between these two tasks. The drawbacks of previous works like the low resolution of instance masks, heavy dependencies of masks on boxes and localization errors of bounding boxes are largely overcome in this work. We argue that object detection and instance segmentation tasks should not be studied separately and hope future work focus on the co-relation between different image perception tasks.

\fontsize{9.0pt}{10.0pt} \selectfont
\bibliographystyle{aaai}
\bibliography{ref.bib}

\end{document}